\pdfoutput=1

\documentclass[11pt]{article}

\usepackage{ACL2023}

\usepackage{times}
\usepackage{latexsym}
\usepackage[T1]{fontenc}

\usepackage[utf8]{inputenc}

\usepackage{microtype}

\usepackage{inconsolata}

\usepackage{graphicx}

\usepackage{multirow}
\usepackage{booktabs}

\usepackage{multibib}
\usepackage{booktabs}
\usepackage{color}
\usepackage{graphicx}
\usepackage{subfigure}
\usepackage{tikz}
\usepackage{pgfplots}
\usepackage{caption}
\usepackage{amsmath}
\usepackage{makecell}
\usepackage{hyperref}
\usepackage{arydshln}
\usepackage{bm}
\usepackage{array}
\usepackage{amsfonts}
\usepackage{times}
\usepackage{latexsym}
\usepackage{color}
\usepackage{tikz}
\usepackage{pgfplots}
\pgfplotsset{compat=1.18}
\usepackage{graphicx}
\usepackage{caption}
\usepackage{amsmath}
\usepackage{makecell}
\usepackage{hyperref}
\usepackage{amssymb}
\usepackage{mathrsfs}
\usepackage{tcolorbox}
\tcbuselibrary{skins, breakable, theorems}
\usepackage{colortbl}

\usepackage{xcolor}
\definecolor{c2}{RGB}{242,242,242}
\definecolor{c1}{RGB}{230,241,243}
\usepackage{xspace}
\newcommand{\method}{PolyVoice\xspace}

\title{\method: Language Models for Speech to Speech Translation}

\author{
{\footnotesize Qianqian Dong\thanks{~~Equal contribution. Working in progress.}~, Zhiying Huang$^\ast$, Qiao Tian, Chen Xu, Tom Ko, Yunlong Zhao, Siyuan Feng, Tang Li, Kexin Wang,}  \\
{\footnotesize \textbf{Xuxin Cheng, Fengpeng Yue, Ye Bai, Xi Chen, Lu Lu, Zejun Ma, Yuping Wang, Mingxuan Wang, Yuxuan Wang}} \\
ByteDance \\
{\tt \{dongqianqian, huangzhiying.92\}@bytedance.com}
}

\begin{document}
\maketitle
\begin{abstract}
We propose \method, a language model-based framework for speech-to-speech translation (S2ST) system.
Our framework consists of
two language models: a translation language model and a speech
synthesis language model.
We use discretized speech units, which are generated in a fully unsupervised way, and thus our framework can be used for unwritten languages.
For the speech synthesis part, we adopt the existing VALL-E X approach and build a unit-based audio language model. 
This grants our framework the ability to preserve the voice characteristics and the speaking style of the original speech.
We examine our system on Chinese $\rightarrow$ English and English $\rightarrow$ Spanish pairs.
Experimental results show that our system can generate speech with high translation quality and audio quality. 
Speech samples are available at \url{https://speechtranslation.github.io/polyvoice}.

\end{abstract}

\section{Introduction}

Speech-to-speech translation (S2ST) is a challenging task as it encounters all the difficulties of automatic speech recognition (ASR), machine translation (MT) and text-to-speech (TTS) synthesis.
Different from conventional cascade approach \cite{Lavie_ICASSP1997,Baldridge_Springer2004,Nakamura_TASLP2006}, the direct approach \cite{Jia_ISCA2019, jia2022translatotron} has the advantages of low latency and simplified pipeline.
Existing direct S2ST approaches can be further classified according to whether the model predicts continuous mel-spectrogram features \cite{dong2022leveraging} or discrete units \cite{Lee-ACL2022}.
Unit-based approach has become more popular due to several reasons: (1) It allows researchers to take advantage of existing NLP modeling techniques by treating acoustic unit as a new language.
(2) It eases the modeling difficulty of emitting spectrogram. 
(3) Units can be generated in a fully unsupervised manner and can cover any unwritten languages.  

There are two kinds of commonly used discretized speech unit: semantic and acoustic units.
Semantic units are usually derived from representations produced by 
speech encoder models like HuBERT \cite{hsu2021hubert}, mHuBERT \cite{lee2021textless} or w2v-BERT \cite{chung2021w2v}. 
They captures the phonetics and semantic content in speech.
Although the making of these units is originally developed to be used as target for training the speech encoder, recently there are attempts to directly use these units as input/output for semantic tasks \cite{meng2022cobert, dongzhang2023}.
Acoustic units can also be referred as codec units. They are originally developed to transmit high-quality speech signal under limited bandwidth.
AudioLM \cite{borsos2022audiolm} is a pioneer work in using language models (LM) for audio generation.
They make use of both kinds of unit and build several LMs with different resolution.
VALL-E \cite{wang2023neural} further extends the AudioLM framework and applies it in TTS.
They successfully demonstrate that the in-context learning capabilities of LM can be similarly replicated in the context of phoneme and codec units. 
In contrast to phoneme units which have to involve supervised training process, both semantic and acoustic units can be generated in a fully unsupervised manner.

Recently, language modeling has made a lot of breakthroughs in NLP.
The success of GPT models \cite{brown2020language, ouyang2022training} is leading the community to a new era. 
Right now, encoder-decoder models are still dominant in speech modeling, where LM-based methods have just begun emerging. 
Thus, we are motivated to investigate the performance of LM-based method in S2ST.
In this paper, we propose a semantic unit-based framework for S2ST system. 
Our framework consists of two LMs: a translation LM and a speech synthesis LM.
The translation LM processes the semantic units of the source language and translates the sequence into semantic units of the target language.
For the speech synthesis part, we adopt the VALL-E X approach \cite{zhang2023speak} for the voice clone ability.
We concatenate the source and target semantic units, as well as the source acoustic units, and feed the whole sequence to the audio LM as a prompt.
The audio LM then predicts the target acoustic units which are converted to a waveform by a unit vocoder.
Experimental results show that our system can generate speech with high translation quality and audio quality. 

We summarise our contribution as follows:

\begin{itemize}
  \item We propose using a decoder-only model to do the direct translation, whereas encoder-decoder model is the dominant structure in previous works.

  \item We build a unit-based audio LM for speech synthesis. Compared to VALL-E X, 
  we use unsupervised discretized unit and can cover unwritten languages.
  
\end{itemize}

The rest of this paper is organized as follows. Section~\ref{sec:related_works} introduces related works in TTS and S2ST. Details of our method are described in Section~\ref{sec:method}. Section~\ref{sec:experiments} introduces our experimental setup. Section~\ref{sec:ablation} presents our ablation study. Finally, we conclude our work in the last section.

\begin{figure*}[!htb]
    \centering
    \includegraphics[scale=0.60]{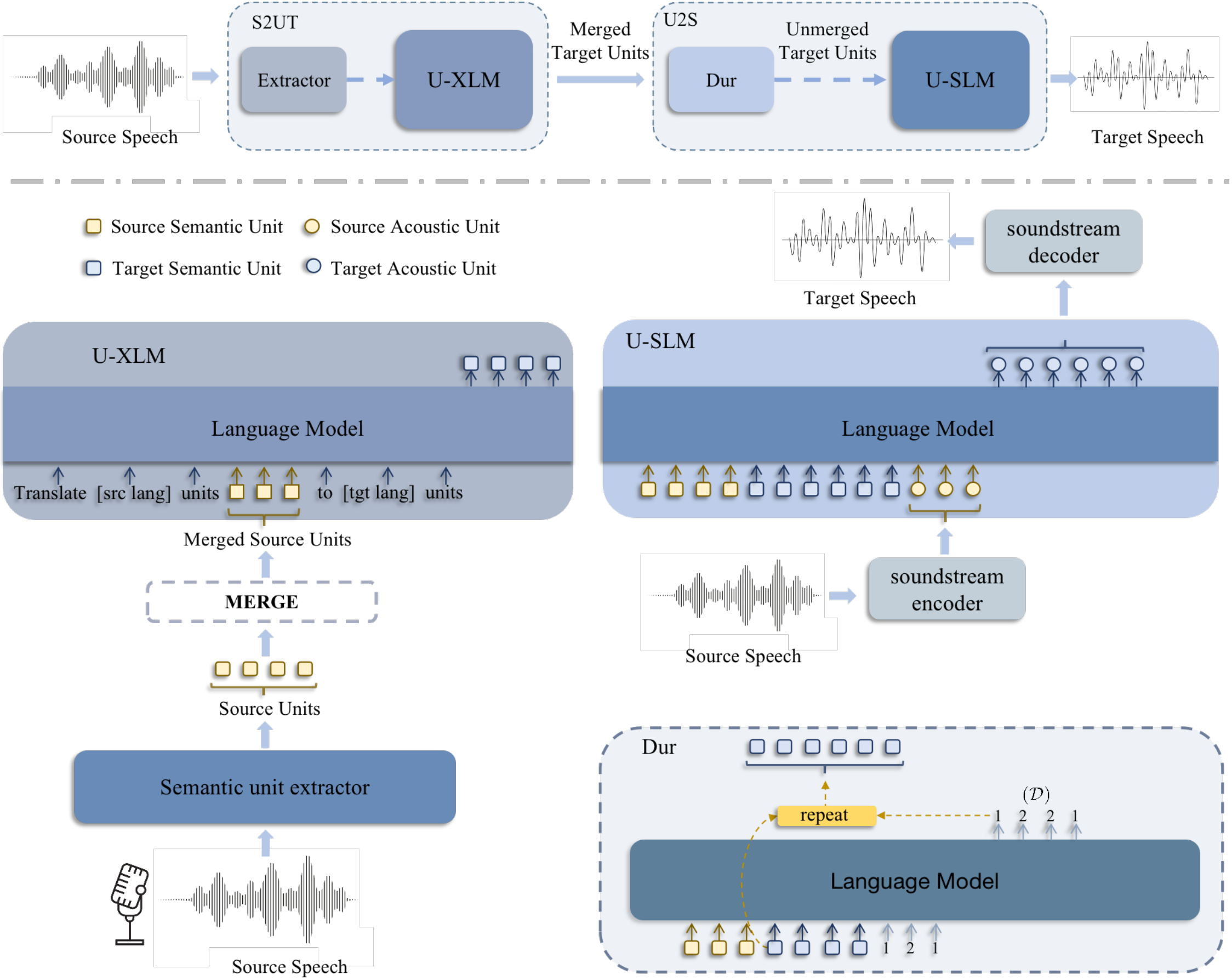}
    \caption{Overview of \method. The framework consists of two LM-based components: a S2UT front-end for translation and a U2S back-end for synthesis.}
    \label{overview}
\end{figure*}

\section{Related Work}
\label{sec:related_works}

\subsection{TTS}

In recent years, neural text-to-speech (TTS) synthesis has achieved significant developments, and the progress of neural network structure makes continuous improvements in the intelligence of synthetic speech \cite{wang17n_interspeech, ren2019fastspeech, kim2021conditional, popov2021grad}.
Because of the requirements of real-world applications, the researchers have attracted a lot of attention to zero-shot multi-speaker TTS and cross-lingual TTS \cite{jia2018transfer,cooper2020zero}.
The multi-speaker TTS  using speaker embedding training on the speaker verification task can generate a similar timbre for a seen speaker. 
However, zero-shot speaker cloning for unseen speakers is still an unsolved problem.
Trained on the large corpus of speech data, VALL-E \cite{wang2023neural} leverages the in-context capability of prefix language modeling to achieve state-of-the-art $($sota$)$ performance for zero-shot speaker cloning.

Cross-lingual TTS aims to build a system that can synthesize speech in a specific language not spoken by the target speaker.
Different embeddings, such as speaker embedding, language embedding, and stress and tone embedding, are utilized in the cross-lingual TTS model to generate high-quality natural and intelligible native speech for native/foreign seen/unseen speakers \cite{liu2019cross}.
Compared with the fixed speaker embedding extracted from the pretrained speaker encoder, a multi-task learning framework has been proposed to enhance cross-lingual speaker similarity by simultaneously training speaker classification \cite{yang2022cross}.
Building upon the prefix language modeling of VALL-E, VALL-E X \cite{zhang2023speak} applies this method to cross-lingual TTS training with bilingual Chinese-English data. 
When presented with source speech, source and target language text prompts, VALL-E X predicts the codec token of the speech in the target language.
Through the in-context capability, the model aims to retain acoustic information from the source speech prompt, such as the acoustic environment, the source language speaker, and their emotion.

\subsection{S2ST}

Speech-to-speech translation \cite{Lavie_ICASSP1997,Baldridge_Springer2004,Nakamura_TASLP2006} aims to develop models capable of generating target language speech from source language speech. 
A naïve system traditionally employs a pipeline \cite{Nakamura_TASLP2006} that sequentially processes the input through automatic speech recognition (ASR) models, machine translation (MT) models, and text-to-speech synthesis (TTS) models.
Recently, end-to-end paradigms \cite{Jia_ISCA2019} have gained popularity in the field of S2ST, as they allow for a single model to perform one or more of the aforementioned tasks, which consequently reduces error propagation and latency. 
Among the various techniques, auxiliary supervision based on textual data has been particularly effective during training \cite{Jia_ISCA2019,Kano_SLT2021_transcoder}. 
However, this approach is not feasible when dealing with unwritten languages.
To address this challenge, discrete units \cite{hsu2021hubert} extracted from the speech are used to replace the target text, and then can be synthesized into the speech \cite{Tjandra_ASRU2019,Zhang_AAAI2021_UWSpeech,Lee-ACL2022}.
Large scale studies have shown the powerful performance in various speech processing tasks \cite{Nguyen_JSTSP2022_Discrete}.

Current research in speech-to-speech translation primarily emphasizes translation quality, with notable improvements observed in automatic evaluation metrics (like BLEU) or human evaluation of naturalness. 
However, there remain two persistent challenges in developing practical systems.
First, these systems are predominantly developed and evaluated on small-scale benchmarks, while real-world scenarios often involve large quantities of labeled data, including ASR, MT, and S2T data.
Even for low-resource or unwritten languages, leveraging unlabeled speech or text can provide valuable information \cite{Lee-ACL2022}. 
Therefore, developing a unified model that jointly utilizes various data types is a critical research goal yet to be achieved.
Second, while not a strict requirement, preserving the source speaker's style during translation is an important aspect of improving user experience \cite{zhang2023speak}. 
However, capturing the unique characteristics of individual speakers is a challenging task. Current approaches, such as speaker embeddings \cite{Jia_ISCA2019} and multi-speaker TTS systems \cite{jia2018transfer}, have made some progress in this direction, but they are still far from for practical requirements.

For the above considerations, We present \method, a versatile framework that can be applied to both written and unwritten language setups. 
\method effectively harnesses diverse data sources within a language model-based framework and preserves the source speaker's style during synthesizing, having the enormous potential in the practical systems.

\section{Method}
\label{sec:method}

\begin{table*}[h]
    \centering
    \resizebox{0.8\textwidth}{!}{
    {
    \begin{tabular}{l}
        \toprule
        \rule{0pt}{10pt}
        \textbf{ASR: [lang]}\\
        \hline
        \vspace{-1mm}
        \begin{tcolorbox}[colback = c2, colframe = c2, width=4.5cm,  halign=left,top=0.5mm, bottom=0.5mm, sharp corners=all]
        {
        \textbf{Data: <unit, text>}
        } 
        \end{tcolorbox}\\
        \quad
        \begin{tcolorbox}[colback = c1, colframe = c1, halign=left,top=0.5mm, bottom=0.5mm, sharp corners=all]
        {
            \textbf{Prompt1}: \textit{Translate \textbf{[lang]} unit `` \textbf{\{unit\}} '' to \textbf{[lang]} text: `` \textbf{\{text\}} ''}  \\
            \textbf{Prompt2}: \textit{Translate \textbf{[lang]} text `` \textbf{\{text\}} '' to \textbf{[lang]} unit: `` \textbf{\{unit\}} ''}  
        }
        \end{tcolorbox}\\
        \hline
        
        \textbf{MT: [src lang] $\rightarrow$ [tgt lang]}  
        \\
        \hline
        \vspace{-1mm}
        \begin{tcolorbox}[colback = c2, colframe = c2, width=6cm,  halign=left,top=0.5mm, bottom=0.5mm, sharp corners=all]
        {
        \textbf{Data: <src\_text, tgt\_text> }
        }
        \end{tcolorbox}\\
        \quad
        \begin{tcolorbox}[colback = c1, colframe = c1, halign=left,top=0.5mm, bottom=0.5mm, sharp corners=all]
        {
        \textbf{Prompt}: \textit{Translate \textbf{[src lang]} text `` \textbf{\{src\_text\}} '' to \textbf{[tgt lang]} text: `` \textbf{\{tgt\_text\}} ''}
        }
        \end{tcolorbox}
        \\
        \hline
        
        \textbf{S2ST: [src lang] $\rightarrow$ [tgt lang]}     \\
        \hline
        \vspace{-1mm}
        \begin{tcolorbox}[colback = c2, colframe = c2, width=9cm,  halign=left,top=0.5mm, bottom=0.5mm, sharp corners=all]
        {
        \textbf{Data: <src\_unit, tgt\_unit, src\_text, tgt\_text>}
        }
        \end{tcolorbox}\\
        \quad
        \begin{tcolorbox}[colback = c1, colframe = c1, halign=left,top=0.5mm, bottom=0.5mm]
        {
        \textbf{Prompt1}: \textit{Translate \textbf{[src lang]} unit `` \textbf{\{src\_unit\}} '' to \textbf{[tgt lang]} unit: `` \textbf{\{tgt\_unit\}} '' }  \\
        \textbf{Prompt2}: \textit{Translate \textbf{[src lang]} unit `` \textbf{\{src\_unit\}} '' to \textbf{[src lang]} text: `` \textbf{\{src\_text\}} '' }  \\
        \textbf{Prompt3}: \textit{Translate \textbf{[src lang]} unit `` \textbf{\{src\_unit\}} '' to \textbf{[tgt lang]} text: `` \textbf{\{tgt\_text\}} ''  } \\
        \textbf{Prompt4}: \textit{Translate \textbf{[src lang]} text `` \textbf{\{src\_text\}} '' to \textbf{[tgt lang]} unit: `` \textbf{\{tgt\_unit\}} '' } \\
        \textbf{Prompt5}: \textit{Translate \textbf{[tgt lang]} text `` \textbf{\{tgt\_text\}} '' to \textbf{[tgt lang]} unit: `` \textbf{\{tgt\_unit\}} '' }
        }
        \end{tcolorbox}\\
        \bottomrule
    \end{tabular}
    }
    }
    \caption{Data construction for U-XLM model by various prompts.}
    \label{prompts}
\end{table*}

We introduce \method, a novel language model-based framework for speech-to-speech translation capable of handling both written and unwritten languages. 
The proposed framework utilizes discrete units, obtained through self-supervised training methods like HuBERT \cite{hsu2021hubert}, as an intermediate representation between source speech and target speech. 
It consists of two parts: a speech-to-unit translation (S2UT) front-end converts the speech in source language into the unit in target language, and a unit-to-speech (U2S) back-end synthesizes speech of translation while preserving the source speaker's style.
Figure \ref{overview} provides an overview of our approach.

\subsection{Speech-to-Unit Translation (S2UT)}

By employing discrete units obtained through self-supervised training, semantically irrelevant information from continuous speech representations is removed, facilitates effective training in an NLP paradigm. 
And S2UT utilizes language model to learn the unit-based cross-lingual generation.

\paragraph{Semantic unit extractor} S2UT first process the raw speech by a semantic unit extractor.
Here we adopt HuBERT, which first encodes the speech by a stack of convolutions and Transformer layers to continuous representations at every 20-ms frame, and then utilizes k-means clustering to discretize the representation to a set of cluster indices $Z={z\_1, \cdots, z\_T}$. 
$T$ is the number of frames and $z\_t \in [K]$, where $K$ is the number of cluster centroids.
Then, we merge the consecutive sequence of duplicate units to compress the sequence length, which reduces the computational costs and help convergence.

\paragraph{Unit-based cross-lingual language model (U-XLM)} 

Over the past few years, the encoder-decoder architecture has emerged as the most prominent paradigm for sequence-to-sequence modeling \cite{sutskever2014sequence}. 
However, recent advances in the GPT family \cite{brown2020language,ouyang2022training} have demonstrated the powerful capability of language modeling by the decoder-only architecture.
This inspires us to develop a unit-based cross-lingual model, that predict the semantic units in target language from the units of source speech by generative language modeling.

We denote the training sample consisting of units of speech in source language and target language as \textit{<src\_unit, tgt\_unit>}.
In the encoder-decoder architecture, the encoder takes the source unit as the input, and the decoder predict the target units.
To enable the cross-lingual unit generation, one can use simple prompts to construct the training samples of natural language from unit pairs, such as: \textit{Translate \textbf{[src lang]} unit `` \textbf{\{src\_unit\}} '' to \textbf{[tgt lang]} unit: `` \textbf{\{tgt\_unit\}}} ''.

\paragraph{Training} 
For training the above U-XLM model, the large scale of data is necessary for competitive performance.
The supervised data, cross-lingual unit pairs, is scarce in real-world scenarios.
Although the auxiliary models can be used to generate the pseudo labels, such as using the TTS model to synthesize the target speech, the direct training of supervised data is expected.

To further address the challenge of data scarcity, previous studies introduce additional loss function into the encoder-decoder architecture through multitask learning \cite{jia2022translatotron,Lee-ACL2022}.
Thanks to language modeling, we adopt a more simple manner to enable the use of diverse data sources like ASR and MT data. 
As shown in Table \ref{prompts}, we slightly modify the prompts to construct training samples for various types of data sources, and then train the model by parameter sharing, simplifying the design of auxiliary objectives.
Unlabeled text and speech can also be used directly in this approach.
In this way, the model implicitly improves the alignment of representation space across speech unit and text.

U-XLM offers several advantages, including the ability to handle both written and unwritten language setups, multilingual modeling capabilities, and the potential for zero-shot prediction by leveraging large amounts of unlabeled data. These features make U-XLM a promising framework for advancing speech-to-speech translation research.

\subsection{Unit-to-speech Synthesis (U2S)}

\paragraph{Unit-to-speech language model (U-SLM)} As shown in Figure \ref{overview}, the U-SLM processes the semantic units predicted by U-XLM and generate the codec units which embed the speaking style of source speaker.
Like VALL-E X, U-SLM includes a autoregressive model and a non-autoregressive model.
Instead of phoneme, discretized semantic units are used in our case.
The unit extractor can be trained in a fully unsupervised manner, which is suitable for  unwritten languages. 

\paragraph{SoundStream codec} We use SoundStream \cite{zeghidour2021soundstream}, a neural audio codec, to compute the embedding of acoustic tokens.
We retrain the SoundStream, whose residual vector quantizer (RVQ) with a hierarchy of 6 vector quantizers and a vocabulary of 1024 symbols.
In our configure, the acoustic tokens is produced at 80Hz for input waveforms at 24 kHz.
This is a 24000 / 80 = 300-fold reduction in the sampling rate.
After the U2S model predict the acoustic tokens represented by the SoundStream codec, the decoder of SoundStream reconstruct them to the waveform.

\paragraph{Duration model} We empirically find that duration information of the discretized unit is very important for the stability of synthesized speech.
In our work, we use a LM to predict the duration.

As shown in Figure \ref{overview}, the merged source semantic unit sequence, merged target semantic unit sequence and the source duration value ($\mathcal{D}$) sequence are concatenated and fed to the duration LM as a prompt.
Then the duration LM predicts the duration value sequence and each target semantic unit will repeat itself accordingly.

\section{Experiments}
\label{sec:experiments}

We evaluate our method on two speech-to-speech benchmark datasets, EMIME~\cite{wester2011emime} and CVSS~\cite{jia2022cvss}. Then, we show the separate results of two components. 

\subsection{Datasets and Preprocessing}

\begin{table}[t!]
    \centering
    \begin{tabular}{lcc}
    \toprule
    Type & Dataset & Size \\
    \midrule
    \multirow{2}*{ASR} & LibriLight (En) & 60K hours   \\
    & In-house (Zh) & 60K hours \\
    \midrule
    MT & In-house & 44M sents \\
    \midrule
    \multirow{2}*{S2S} & GigaSpeech & 10K hours \\
    & WenetSpeech & 10K hours \\
    \bottomrule
    \end{tabular}
    \caption{Training data of U-XLM model.}
    \label{data_stat}
\end{table}

\begin{table*}[htbp!]
\resizebox{\textwidth}{!}{
\begin{tabular}{@{}lccccc@{}}
\toprule
\multirow{2}{*}{}   & \multicolumn{3}{c}{ASV $\uparrow$}    & \multirow{2}{*}{ASR-BLEU $\uparrow$ } & \multicolumn{1}{r}{\multirow{2}{*}{Naturalness $\uparrow$ }} \\ 
\cmidrule(lr){2-4}
& tgt vs. src     & \multicolumn{1}{l}{hyp vs. src} & hyp vs. tgt &  & \multicolumn{1}{r}{}    \\ 
\midrule
\multirow{2}{*}{\begin{tabular}[c]{@{}l@{}}Cascade (VALL-E X  paper)\\ ~~~~+ w/ oracle target text\end{tabular}} 
 & \multirow{4}{*}{0.58} & 0.28   & 0.27   & 27.49 & 3.44     \\
 &   & 0.28   & 0.29   & 80.30 & 3.43     \\ 
\cmidrule(r){1-1}
\multirow{2}{*}{\begin{tabular}[c]{@{}l@{}}VALL-E X (VALL-E X paper)\\ ~~~~+ w/ oracle target text\end{tabular}} 
 &   & 0.37   & 0.37   & 30.66 & 3.54     \\
 &   & 0.39   & 0.38   & 86.78 & 3.54     \\ 
\cmidrule(r){1-1}
\multirow{3}{*}{
\begin{tabular}[c]{@{}l@{}}
    S2UT \\ 
    \method (S2UT + U2S) \\
    ~~~~~+ w/ oracle target semantic unit 
    \end{tabular}
} 
 &  \multirow{3}{*}{0.59} & 0.06   & 0.08   & 29.30  & 3.35     \\ 
 &  & 0.38   & 0.38   & 29.40  & 4.10     \\ 
 &  & 0.42   & 0.48   & 76.10  & 3.92     \\ 
\bottomrule
\end{tabular}
}
\caption{S2ST results on Chinese-English EMIME dataset.}
\label{results_s2st}
\end{table*}

\subsubsection{S2UT}

\paragraph{Semantic token}
U-XLM is trained by cross-lingual unit data, which is extracted from the audio by HuBERT \cite{hsu2021hubert} models. For Chinese audio, we utilize an open-source model based on WenetSpeech Chinese speech~\footnote{\href{https://github.com/TencentGameMate/chinese\_speech\_pretrain}{https://github.com/TencentGameMate/chinese\_speech\_\\pretrain}}. 
For English and Spanish audio, we use an open-source multilingual model (English, Spanish and French)~\footnote{\href{https://github.com/facebookresearch/fairseq/blob/main/examples/speech\_to\_speech/docs/textless\_s2st\_real\_data.md}{https://github.com/facebookresearch/fairseq/blob/main/\\examples/speech\_to\_speech/docs/textless\_s2st\_real\_data.md}}.
The cluster centroids of k-mean algorithm for two models are 500 and 1,000, respectively.

\paragraph{Vocabulary}
To address the out-of-vocabulary problem and enable parameter sharing across languages, we utilize byte-level subword units \footnote{\href{https://github.com/huggingface/tokenizers}{https://github.com/huggingface/tokenizers}} that decompose each character into byte-sized pieces, achieves a vocabulary size of 56,407 (including 1,500 cluster centroids).

\paragraph{Datasets}
Considered that the paired speech-to-speech (S2S) data is scarcity, we synthesize the pseudo data from the ASR data utilizing in-house MT and TTS systems.
In addition, various types of data resources provide better learning of the U-XLM model, like large-scale ASR and MT data.
The detailed statistics are shown in Table \ref{data_stat}. 

The S2S data is sourced from WenetSpeech \cite{Zhang_ISCA2022_Wenetspeech} and GigaSpeech \cite{Chen_ISCA2021_Gigaspeech}.
WenetSpeech is a Chinese ASR dataset with over 10,000 hours of speech data collected from YouTube. 
And we utilize a subset of 10,000 hours of GigaSpeech \cite{Chen_ISCA2021_Gigaspeech}, an English ASR dataset collected from audiobooks, podcasts, and YouTube.

Then we scale up the training data using specific prompts for various types of dataset.
We utilize the LibriLight \cite{Kahn_ICASSP2020_Librilight} and the in-house ASR datasets.
LibriLight is an unlabeled English speech dataset containing about 60,000 hours of speech.
Since LibriLight has many long audios, we segment and recognize the audio based on the method of voice active detection (VAD) and in-house ASR system, generating the audio length ranging from 0.5 to 25s, and the average length is 7s.
In-house ASR dataset is a Chinese ASR dataset with 60,000 hours of speech.
We also use the in-house Chinese-English MT dataset consisting of 44M sentence pairs.

\subsubsection{U2S}

The U-SLM is trained on the large open-source bilingual speech data, i.e., WenetSpeech \cite{Zhang_ISCA2022_Wenetspeech} and LibriLight \cite{Kahn_ICASSP2020_Librilight}.
The Librilight is handled in the same way as U-XLM.
WenetSpeech keeps the original data length unchanged, the audio length ranges from 0.5 to 20s, and the average length is 2.5s.
In addition, we used an additional 250h internal Chinese TTS data and 400h internal English TTS data.

\subsection{Evaluation}

To measure the performance of our system, we evaluate both the translation quality and the speech quality.

\paragraph{Translation Quality} 
Following the previous setups, we recognize the speech output by an in-house ASR system to compute BLEU scores (ASR-BLEU) for S2ST results.

\paragraph{Speech Quality}
The speech quality is evaluated by multiple metrics. 
The capability of voice clone is measured by the speaker similarity (ASV-Score), which is calculated by an ASV model~\footnote{\href{https://github.com/Sanyuan-Chen/UniSpeech/tree/t-schen/asv\_eval/downstreams/speaker\_verification\#example-2}{https://github.com/Sanyuan-Chen/UniSpeech/tree/t-sch\\en/asv\_eval/downstreams/speaker\_verification\#example-2}} to determine whether the synthesized speech is from the same speaker as the ground-truth speech.
The naturalness of the speech output is evaluated by the automatic metric using \texttt{NISQA}~\footnote{\href{https://github.com/gabrielmittag/NISQA}{https://github.com/gabrielmittag/NISQA}}. And the pronunciation accuracy is evaluated using WER scores (ASR-WER) with a ASR model based on \texttt{hubert-large}~\footnote{\href{https://huggingface.co/facebook/hubert-large-ls960-ft}{https://huggingface.co/facebook/hubert-large-ls960-ft}}.

\subsection{Model Settings}

\begin{table*}[htbp!] 

\centering
\begin{tabular}{lccc}
\toprule
CVSS    & \multicolumn{1}{l}{ASV $\uparrow$} & \multicolumn{1}{l}{\begin{tabular}[c]{@{}l@{}}BLEU $\uparrow$ \end{tabular}} & \multicolumn{1}{l}{\begin{tabular}[c]{@{}l@{}}Naturalness$\uparrow$\end{tabular}} \\ 
\hline
Ground-truth & 0.19 & 89.3    & 3.54      \\
\method     & 0.34    & 18.3       & 3.60    \\
~~~~+ w/ oracle target unit & 0.28 & 70.8    & 3.69      \\ 
\bottomrule
\end{tabular}
\caption{Results on the English-Spanish CVSS dataset. We train the model with paired speech-to-speech datasets expanded from GigaSpeech without any text information. BLEU means ASR-BLEU, target unit means oracle Spanish unit.}
\label{results_unwritten}
\end{table*}

\subsubsection{S2UT}

In the S2UT front-end, U-XLM's model architecture is a unidirectional Transformer decoder consisting of 48 layers with hidden size 1600, feed-forward network (FFN) size 6400, and 25 attention heads. The total parameters are 1.6 B.
U-XLM is trained on 8/32 NVIDIA TESLA A100 80GB GPUs with a batch size of 3072 tokens per GPU for 500k steps.

\subsubsection{U2S}

In the U2S back-end, the U-SLM consists of 12 transformer layers. Each of these layers comprises 16 attention heads, an attention dimension of 1024, and an FFN dimension of 4096 in both the autoregressive (AR) model and non-autoregressive (NAR) model.
We train the models using 8 NVIDIA TESLA A100 80GB GPUs, with a batch size of 8 utterances per GPU for 800k steps.
Training for all steps takes about 5 days.

\subsection{Results and Analysis}

\subsubsection{S2ST Results}

Table \ref{results_s2st} summarizes the overall performance of our method for S2ST.
We conduct experiments on the EMIME dataset to enable direct comparisons with the most similar work VALL-E X.
The cascade system treats S2ST as a pipeline of running an ASR model, an MT model, and a multi-speaker YourTTS model sequentially. During the synthesis process, speaker information is integrated using speaker embeddings.

We first evaluate the capability to preserve the voice of the source speaker in the output speech, using the ASV score. 
We calculate speaker similarity between the source speech, target speech, and synthesized speech. 
We run the U-XLM alone, where speech is synthesized by a Unit-based vocoder\footnote{\href{https://github.com/facebookresearch/fairseq/blob/main/examples/speech\_to\_speech/docs/textless\_s2st\_real\_data.md}{https://github.com/facebookresearch/fairseq/blob/main/\\examples/speech\_to\_speech/docs/textless\_s2st\_real\_data.md}}.
Due to the lack of explicit modeling of speaker characteristics, it produces particularly low ASV scores.
Both the VALL-E X and \method systems, which adopt in-context learning, show superior performance over the speaker embedding.
Notably, our method demonstrates better voice cloning capabilities when ground-truth target information was available.

\method achieves a slight degraded translation quality (ASR-BLEU) but a remarkable improvement in speech quality (naturalness) compared with VALL-E X.
When taking the ground-truth target information as input, \method is inferior to VALL-E X with a large gap of about 10 BLEU points, while the naturalness improves significantly.
The semantic units are extracted from the speech by unsupervised learning, which inevitably introduces errors.
Although units are considered ``semantic'' tokens,  they still preserve some acoustic information.
Therefore, unit-based modeling leads to better speech quality but worse translation quality.
In contrast, phonemes obtained from the text ensure semantic correctness but lost the acoustic information.
Therefore, we believe that units have more potential, even if the current performance is slightly degraded.
And future work can focus on enhancing the extraction of semantic information to improve translation quality.

Interestingly, \method achieves better naturalness using the predicted units.
We speculate that this is due to the language model's output having better fluency.
U-XLM learns the speech distribution over the large scale of unit data, and tends to generate more natural sequences of units.
However, this may interfere with the accuracy of the translation.
We will explore this issue in the future.

\subsubsection{Unwritten Language Scenario}
We examine our proposed framework in the case where the source is a written language and the target is a unwritten language.
In our setup, we train and evaluate an English$\rightarrow$Spanish S2ST system without the use of any Spanish text transcript.
Table \ref{results_unwritten} summarizes the results.
The ASR-BLEU (18.3) indicates that the Spanish speech generated by our system is semantically understandable.
This demonstrates the ability of our S2ST system for the unwritten languages.

\section{Ablation Study}
\label{sec:ablation}

\subsection{Decoder-only vs. Encoder-Decoder}

\begin{table}[tbp!]
    \centering
    \begin{tabular}{lc}
    \toprule
    Arch & ASR-BLEU \\
    \midrule
    Encoder-Decoder & 16.8 \\
    ~~~~ + w/ U2S & 18.7 \\
    Decoder-only & 20.7 \\
    ~~~~ + w/ U2S & 22.0 \\
    \bottomrule
    \end{tabular}
    \caption{Performance with different architectures.}
    \label{results_arch}
\end{table}

\begin{table*}[htbp!]
    \centering
    \begin{tabular}{lccccc}
    \toprule
    Task & S2ST (BLEU $\uparrow$) & ASR (CER $\downarrow$) & ST (BLEU $\uparrow$) & MT (BLEU $\uparrow$) & TTS (WER $\downarrow$)\\
    \midrule
    S2S & 22.2 &-&-&-&-\\
    ~~~~+ MTL & 29.4 &4.46 & 30.8 & 33.81 & 6.99 \\
    \bottomrule
    \end{tabular}
    \caption{The performance of multiple tasks on EMIME dataset. Here are the explanations for each task. S2ST: Chinese speech to English speech; ASR: Chinese speech to Chinese text; ST: Chinese speech to English text; MT: Chinese text to English text; TTS: English text to English speech.}
    \label{results_mtl}
\end{table*}

\begin{table*}[htbp!]
\centering
\begin{tabular}{lccc}
\toprule
Methods  & WER $\downarrow$ & ASV $\uparrow$ & Naturalness $\uparrow$ \\ 
\midrule
VALL-E X (paper)   & 4.07    & 0.36 & 3.54      \\
U2S   & 6.40    & 0.38 & 3.98           \\
~~~~+ w/o semantic2dur & 31.93   & 0.37 & 3.81  \\
~~~~+ w/ mHuBERT\_zh\_en & 4.76    & 0.37 & 3.81 \\
\bottomrule
\end{tabular}
\caption{Evaluation of the speech synthesizers.}
\label{TTS_result}
\end{table*}

Empirical studies in the field of natural language processing have revealed that the full potential of the decoder-only approach can be realized through the use of large model sizes and expansive datasets. 
As pioneers in exploring the application of language models to S2ST, we present a fair comparison of the two architectures in Table \ref{results_arch}.

Two models are trained with same training data.
Interestingly, the decoder-only model yields a remarkable improvement of 3.9 BLEU points over the encoder-decoder counterpart\footnote{We train the encoder-decoder architecture using the code: \href{https://github.com/facebookresearch/fairseq/blob/main/examples/speech\_to\_speech/docs/direct\_s2st\_discrete\_units.md}{https://github.com/facebookresearch/fairseq/blob/main/examp\\les/speech\_to\_speech/docs/direct\_s2st\_discrete\_units.md}.}.
When we synthesize the speech by U2S instead of vocoder, the performance gap is reduced, highlighting the robustness of our U2S back-end.

\subsection{Multi-task Training}

As discussed in Section~\ref{sec:method}, the language modeling enables the direct training over the diverse data sources utilizing specific prompts.
In this way, we combine additional large scale ASR and MT data to fully explore the potential of our method.

As shown in Table~\ref{results_mtl}, U-XLM achieves promising performance for multiple tasks involved (including S2ST, ASR, ST, MT, and TTS) under the expanded data setting, which verifies the capability of the general modeling in the decoder-only architecture.
In the traditional paradigm, we need to design the complex manner to combine multi-task learning, but language modeling only modify the prompt to construct the training data.

\subsection{Semantic Unit and Duration Model}

Table \ref{TTS_result} shows the resynthesis performance of different speech synthesizers. 
Our TTS obtains better performance in both ASV and naturalness. 
We attribute the increase of WER to the difference in amount of semantic information carried by phonemes and unsupervised units.
This is consistent with the observation reported in the work of mHuBERT and AudioLM.

If we remove the duration model from the U2S, the WER increases dramatically.
Our guess is that the unit itself do not contain as many duration information as the phonemes. Therefore the duration model is essential when using unsupervised units.

We further train our own multilingual HuBERT model (mHuBERT\_zh\_en) with a combination of Chinese and English data. The model size is the same as the HuBERT-large model in \cite{hsu2021hubert}.
We find that the WER improves when we use the semantic units generated from mHuBERT\_zh\_en. Thus, we believe that a larger model may generate better semantic units.
We do not use mHuBERT\_zh\_en in our S2ST experiment because we need the mHuBERT in \cite{lee2021textless} to run the English->Spanish experiment.
The benefit of using mHuBERT\_zh\_en to the overall S2ST is left for future work.

\section{Conclusion and Future Work}
In this paper, we propose a semantic unit-based framework for S2ST. Our framework consists of two LMs: a translation LM (U-XLM) and a speech synthesis LM (U-SLM). We show that our unit-based S2ST system performs better than existing systems in terms of ASR-BLEU, ASV and naturalness. Furthermore, we demonstrate the system ability in unwritten language scenario without any use of the Spanish text transcript.
As our system performance is highly related to the quality of the semantic units, future work will investigate 
the way to generate a better set of discrete units.
Also, we plan to investigate how the performance can be further improved by using much larger model.

\clearpage

\bibliography{custom}

\begin{thebibliography}{36}
\expandafter\ifx\csname natexlab\endcsname\relax\def\natexlab#1{#1}\fi

\bibitem[{Baldridge(2004)}]{Baldridge_Springer2004}
Jason Baldridge. 2004.
\newblock \href {https://doi.org/10.1017/S1351324904233435} {\emph{Verbmobil:
  Foundations of Speech-to-Speech Translation}, by wolfgang wahlster (editor).
  springer, 2000. {ISBN} 3-540-67783-6. price {\textsterling}44.50 (hardback).
  xii+679 pages}.
\newblock \emph{Nat. Lang. Eng.}, 10(2):200--204.

\bibitem[{Borsos et~al.(2022)Borsos, Marinier, Vincent, Kharitonov, Pietquin,
  Sharifi, Teboul, Grangier, Tagliasacchi, and Zeghidour}]{borsos2022audiolm}
Zal{\'a}n Borsos, Rapha{\"e}l Marinier, Damien Vincent, Eugene Kharitonov,
  Olivier Pietquin, Matt Sharifi, Olivier Teboul, David Grangier, Marco
  Tagliasacchi, and Neil Zeghidour. 2022.
\newblock Audiolm: a language modeling approach to audio generation.
\newblock \emph{arXiv preprint arXiv:2209.03143}.

\bibitem[{Brown et~al.(2020)Brown, Mann, Ryder, Subbiah, Kaplan, Dhariwal,
  Neelakantan, Shyam, Sastry, Askell et~al.}]{brown2020language}
Tom Brown, Benjamin Mann, Nick Ryder, Melanie Subbiah, Jared~D Kaplan, Prafulla
  Dhariwal, Arvind Neelakantan, Pranav Shyam, Girish Sastry, Amanda Askell,
  et~al. 2020.
\newblock Language models are few-shot learners.
\newblock \emph{Advances in neural information processing systems},
  33:1877--1901.

\bibitem[{Chen et~al.(2021)Chen, Chai, Wang, Du, Zhang, Weng, Su, Povey, Trmal,
  Zhang, Jin, Khudanpur, Watanabe, Zhao, Zou, Li, Yao, Wang, You, and
  Yan}]{Chen_ISCA2021_Gigaspeech}
Guoguo Chen, Shuzhou Chai, Guan{-}Bo Wang, Jiayu Du, Wei{-}Qiang Zhang, Chao
  Weng, Dan Su, Daniel Povey, Jan Trmal, Junbo Zhang, Mingjie Jin, Sanjeev
  Khudanpur, Shinji Watanabe, Shuaijiang Zhao, Wei Zou, Xiangang Li, Xuchen
  Yao, Yongqing Wang, Zhao You, and Zhiyong Yan. 2021.
\newblock \href {https://doi.org/10.21437/Interspeech.2021-1965} {Gigaspeech:
  An evolving, multi-domain {ASR} corpus with 10, 000 hours of transcribed
  audio}.
\newblock In \emph{Interspeech 2021, 22nd Annual Conference of the
  International Speech Communication Association, Brno, Czechia, 30 August - 3
  September 2021}, pages 3670--3674. {ISCA}.

\bibitem[{Chung et~al.(2021)Chung, Zhang, Han, Chiu, Qin, Pang, and
  Wu}]{chung2021w2v}
Yu-An Chung, Yu~Zhang, Wei Han, Chung-Cheng Chiu, James Qin, Ruoming Pang, and
  Yonghui Wu. 2021.
\newblock W2v-bert: Combining contrastive learning and masked language modeling
  for self-supervised speech pre-training.
\newblock In \emph{2021 IEEE Automatic Speech Recognition and Understanding
  Workshop (ASRU)}, pages 244--250. IEEE.

\bibitem[{Cooper et~al.(2020)Cooper, Lai, Yasuda, Fang, Wang, Chen, and
  Yamagishi}]{cooper2020zero}
Erica Cooper, Cheng-I Lai, Yusuke Yasuda, Fuming Fang, Xin Wang, Nanxin Chen,
  and Junichi Yamagishi. 2020.
\newblock Zero-shot multi-speaker text-to-speech with state-of-the-art neural
  speaker embeddings.
\newblock In \emph{ICASSP 2020-2020 IEEE International Conference on Acoustics,
  Speech and Signal Processing (ICASSP)}, pages 6184--6188. IEEE.

\bibitem[{Dong et~al.(2022)Dong, Yue, Ko, Wang, Bai, and
  Zhang}]{dong2022leveraging}
Qianqian Dong, Fengpeng Yue, Tom Ko, Mingxuan Wang, Qibing Bai, and Yu~Zhang.
  2022.
\newblock Leveraging pseudo-labeled data to improve direct speech-to-speech
  translation.

\bibitem[{Hsu et~al.(2021)Hsu, Bolte, Tsai, Lakhotia, Salakhutdinov, and
  Mohamed}]{hsu2021hubert}
Wei-Ning Hsu, Benjamin Bolte, Yao-Hung~Hubert Tsai, Kushal Lakhotia, Ruslan
  Salakhutdinov, and Abdelrahman Mohamed. 2021.
\newblock Hubert: Self-supervised speech representation learning by masked
  prediction of hidden units.
\newblock \emph{IEEE/ACM Transactions on Audio, Speech, and Language
  Processing}, 29:3451--3460.

\bibitem[{Jia et~al.(2022{\natexlab{a}})Jia, Ramanovich, Remez, and
  Pomerantz}]{jia2022translatotron}
Ye~Jia, Michelle~Tadmor Ramanovich, Tal Remez, and Roi Pomerantz.
  2022{\natexlab{a}}.
\newblock Translatotron 2: High-quality direct speech-to-speech translation
  with voice preservation.
\newblock In \emph{International Conference on Machine Learning}, pages
  10120--10134. PMLR.

\bibitem[{Jia et~al.(2022{\natexlab{b}})Jia, Ramanovich, Wang, and
  Zen}]{jia2022cvss}
Ye~Jia, Michelle~Tadmor Ramanovich, Quan Wang, and Heiga Zen.
  2022{\natexlab{b}}.
\newblock Cvss corpus and massively multilingual speech-to-speech translation.
\newblock \emph{arXiv preprint arXiv:2201.03713}.

\bibitem[{Jia et~al.(2019)Jia, Weiss, Biadsy, Macherey, Johnson, Chen, and
  Wu}]{Jia_ISCA2019}
Ye~Jia, Ron~J. Weiss, Fadi Biadsy, Wolfgang Macherey, Melvin Johnson, Zhifeng
  Chen, and Yonghui Wu. 2019.
\newblock \href {https://doi.org/10.21437/Interspeech.2019-1951} {Direct
  speech-to-speech translation with a sequence-to-sequence model}.
\newblock In \emph{Interspeech 2019, 20th Annual Conference of the
  International Speech Communication Association, Graz, Austria, 15-19
  September 2019}, pages 1123--1127. {ISCA}.

\bibitem[{Jia et~al.(2018)Jia, Zhang, Weiss, Wang, Shen, Ren, Nguyen, Pang,
  Lopez~Moreno, Wu et~al.}]{jia2018transfer}
Ye~Jia, Yu~Zhang, Ron Weiss, Quan Wang, Jonathan Shen, Fei Ren, Patrick Nguyen,
  Ruoming Pang, Ignacio Lopez~Moreno, Yonghui Wu, et~al. 2018.
\newblock Transfer learning from speaker verification to multispeaker
  text-to-speech synthesis.
\newblock \emph{Advances in neural information processing systems}, 31.

\bibitem[{Kahn et~al.(2020)Kahn, Rivi{\`{e}}re, Zheng, Kharitonov, Xu,
  Mazar{\'{e}}, Karadayi, Liptchinsky, Collobert, Fuegen, Likhomanenko,
  Synnaeve, Joulin, Mohamed, and Dupoux}]{Kahn_ICASSP2020_Librilight}
Jacob Kahn, Morgane Rivi{\`{e}}re, Weiyi Zheng, Evgeny Kharitonov, Qiantong Xu,
  Pierre{-}Emmanuel Mazar{\'{e}}, Julien Karadayi, Vitaliy Liptchinsky, Ronan
  Collobert, Christian Fuegen, Tatiana Likhomanenko, Gabriel Synnaeve, Armand
  Joulin, Abdelrahman Mohamed, and Emmanuel Dupoux. 2020.
\newblock \href {https://doi.org/10.1109/ICASSP40776.2020.9052942}
  {Libri-light: {A} benchmark for {ASR} with limited or no supervision}.
\newblock In \emph{2020 {IEEE} International Conference on Acoustics, Speech
  and Signal Processing, {ICASSP} 2020, Barcelona, Spain, May 4-8, 2020}, pages
  7669--7673. {IEEE}.

\bibitem[{Kano et~al.(2021)Kano, Sakti, and Nakamura}]{Kano_SLT2021_transcoder}
Takatomo Kano, Sakriani Sakti, and Satoshi Nakamura. 2021.
\newblock \href {https://doi.org/10.1109/SLT48900.2021.9383496}
  {Transformer-based direct speech-to-speech translation with transcoder}.
\newblock In \emph{{IEEE} Spoken Language Technology Workshop, {SLT} 2021,
  Shenzhen, China, January 19-22, 2021}, pages 958--965. {IEEE}.

\bibitem[{Kim et~al.(2021)Kim, Kong, and Son}]{kim2021conditional}
Jaehyeon Kim, Jungil Kong, and Juhee Son. 2021.
\newblock Conditional variational autoencoder with adversarial learning for
  end-to-end text-to-speech.
\newblock In \emph{International Conference on Machine Learning}, pages
  5530--5540. PMLR.

\bibitem[{Lavie et~al.(1997)Lavie, Waibel, Levin, Finke, Gates, Gavald{\`{a}},
  Zeppenfeld, and Zhan}]{Lavie_ICASSP1997}
Alon Lavie, Alex Waibel, Lori~S. Levin, Michael Finke, Donna Gates, Marsal
  Gavald{\`{a}}, Torsten Zeppenfeld, and Puming Zhan. 1997.
\newblock \href {https://doi.org/10.1109/ICASSP.1997.599557} {Janus-iii:
  speech-to-speech translation in multiple languages}.
\newblock In \emph{1997 {IEEE} International Conference on Acoustics, Speech,
  and Signal Processing, {ICASSP} '97, Munich, Germany, April 21-24, 1997},
  pages 99--102. {IEEE} Computer Society.

\bibitem[{Lee et~al.(2022)Lee, Chen, Wang, Gu, Popuri, Ma, Polyak, Adi, He,
  Tang, Pino, and Hsu}]{Lee-ACL2022}
Ann Lee, Peng{-}Jen Chen, Changhan Wang, Jiatao Gu, Sravya Popuri, Xutai Ma,
  Adam Polyak, Yossi Adi, Qing He, Yun Tang, Juan Pino, and Wei{-}Ning Hsu.
  2022.
\newblock \href {https://doi.org/10.18653/v1/2022.acl-long.235} {Direct
  speech-to-speech translation with discrete units}.
\newblock In \emph{Proceedings of the 60th Annual Meeting of the Association
  for Computational Linguistics (Volume 1: Long Papers), {ACL} 2022, Dublin,
  Ireland, May 22-27, 2022}, pages 3327--3339. Association for Computational
  Linguistics.

\bibitem[{Lee et~al.(2021)Lee, Gong, Duquenne, Schwenk, Chen, Wang, Popuri,
  Adi, Pino, Gu et~al.}]{lee2021textless}
Ann Lee, Hongyu Gong, Paul-Ambroise Duquenne, Holger Schwenk, Peng-Jen Chen,
  Changhan Wang, Sravya Popuri, Yossi Adi, Juan Pino, Jiatao Gu, et~al. 2021.
\newblock Textless speech-to-speech translation on real data.
\newblock \emph{arXiv preprint arXiv:2112.08352}.

\bibitem[{Liu and Mak(2019)}]{liu2019cross}
Zhaoyu Liu and Brian Mak. 2019.
\newblock Cross-lingual multi-speaker text-to-speech synthesis for voice
  cloning without using parallel corpus for unseen speakers.
\newblock \emph{arXiv preprint arXiv:1911.11601}.

\bibitem[{Meng et~al.()Meng, Ao, Ko, Wang, and Li}]{meng2022cobert}
Chutong Meng, Junyi Ao, Tom Ko, Mingxuan Wang, and Haizhou Li.
\newblock Cobert: Self-supervised speech representation learning through code
  representation learning.
\newblock In \emph{Interspeech 2023}.

\bibitem[{Nakamura et~al.(2006)Nakamura, Markov, Nakaiwa, Kikui, Kawai,
  Jitsuhiro, Zhang, Yamamoto, Sumita, and Yamamoto}]{Nakamura_TASLP2006}
Satoshi Nakamura, Konstantin Markov, Hiromi Nakaiwa, Gen{-}ichiro Kikui,
  Hisashi Kawai, Takatoshi Jitsuhiro, Jinsong Zhang, Hirofumi Yamamoto,
  Eiichiro Sumita, and Seiichi Yamamoto. 2006.
\newblock \href {https://doi.org/10.1109/TSA.2005.860774} {The {ATR}
  multilingual speech-to-speech translation system}.
\newblock \emph{{IEEE} Trans. Speech Audio Process.}, 14(2):365--376.

\bibitem[{Nguyen et~al.(2022)Nguyen, Sagot, and
  Dupoux}]{Nguyen_JSTSP2022_Discrete}
Tu~Anh Nguyen, Beno{\^{\i}}t Sagot, and Emmanuel Dupoux. 2022.
\newblock \href {https://doi.org/10.1109/JSTSP.2022.3200909} {Are discrete
  units necessary for spoken language modeling?}
\newblock \emph{{IEEE} J. Sel. Top. Signal Process.}, 16(6):1415--1423.

\bibitem[{Ouyang et~al.(2022)Ouyang, Wu, Jiang, Almeida, Wainwright, Mishkin,
  Zhang, Agarwal, Slama, Ray et~al.}]{ouyang2022training}
Long Ouyang, Jeffrey Wu, Xu~Jiang, Diogo Almeida, Carroll Wainwright, Pamela
  Mishkin, Chong Zhang, Sandhini Agarwal, Katarina Slama, Alex Ray, et~al.
  2022.
\newblock Training language models to follow instructions with human feedback.
\newblock \emph{Advances in Neural Information Processing Systems},
  35:27730--27744.

\bibitem[{Popov et~al.(2021)Popov, Vovk, Gogoryan, Sadekova, and
  Kudinov}]{popov2021grad}
Vadim Popov, Ivan Vovk, Vladimir Gogoryan, Tasnima Sadekova, and Mikhail
  Kudinov. 2021.
\newblock Grad-tts: A diffusion probabilistic model for text-to-speech.
\newblock In \emph{International Conference on Machine Learning}, pages
  8599--8608. PMLR.

\bibitem[{Ren et~al.(2019)Ren, Ruan, Tan, Qin, Zhao, Zhao, and
  Liu}]{ren2019fastspeech}
Yi~Ren, Yangjun Ruan, Xu~Tan, Tao Qin, Sheng Zhao, Zhou Zhao, and Tie-Yan Liu.
  2019.
\newblock Fastspeech: Fast, robust and controllable text to speech.
\newblock \emph{Advances in neural information processing systems}, 32.

\bibitem[{Sutskever et~al.(2014)Sutskever, Vinyals, and
  Le}]{sutskever2014sequence}
Ilya Sutskever, Oriol Vinyals, and Quoc~V. Le. 2014.
\newblock \href
  {https://proceedings.neurips.cc/paper/2014/hash/a14ac55a4f27472c5d894ec1c3c743d2-Abstract.html}
  {Sequence to sequence learning with neural networks}.
\newblock In \emph{Advances in Neural Information Processing Systems 27: Annual
  Conference on Neural Information Processing Systems 2014, December 8-13 2014,
  Montreal, Quebec, Canada}, pages 3104--3112.

\bibitem[{Tjandra et~al.(2019)Tjandra, Sakti, and Nakamura}]{Tjandra_ASRU2019}
Andros Tjandra, Sakriani Sakti, and Satoshi Nakamura. 2019.
\newblock \href {https://doi.org/10.1109/ASRU46091.2019.9003853}
  {Speech-to-speech translation between untranscribed unknown languages}.
\newblock In \emph{{IEEE} Automatic Speech Recognition and Understanding
  Workshop, {ASRU} 2019, Singapore, December 14-18, 2019}, pages 593--600.
  {IEEE}.

\bibitem[{Wang et~al.(2023)Wang, Chen, Wu, Zhang, Zhou, Liu, Chen, Liu, Wang,
  Li et~al.}]{wang2023neural}
Chengyi Wang, Sanyuan Chen, Yu~Wu, Ziqiang Zhang, Long Zhou, Shujie Liu, Zhuo
  Chen, Yanqing Liu, Huaming Wang, Jinyu Li, et~al. 2023.
\newblock Neural codec language models are zero-shot text to speech
  synthesizers.
\newblock \emph{arXiv preprint arXiv:2301.02111}.

\bibitem[{Wang et~al.(2017)Wang, Skerry-Ryan, Stanton, Wu, Weiss, Jaitly, Yang,
  Xiao, Chen, Bengio, Le, Agiomyrgiannakis, Clark, and
  Saurous}]{wang17n_interspeech}
Yuxuan Wang, R.J. Skerry-Ryan, Daisy Stanton, Yonghui Wu, Ron~J. Weiss, Navdeep
  Jaitly, Zongheng Yang, Ying Xiao, Zhifeng Chen, Samy Bengio, Quoc Le, Yannis
  Agiomyrgiannakis, Rob Clark, and Rif~A. Saurous. 2017.
\newblock \href {https://doi.org/10.21437/Interspeech.2017-1452} {{Tacotron:
  Towards End-to-End Speech Synthesis}}.
\newblock In \emph{Proc. Interspeech 2017}, pages 4006--4010.

\bibitem[{Wester and Liang(2011)}]{wester2011emime}
Mirjam Wester and Hui Liang. 2011.
\newblock The emime mandarin bilingual database.
\newblock Technical report, The University of Edinburgh.

\bibitem[{Yang and He(2022)}]{yang2022cross}
Jingzhou Yang and Lei He. 2022.
\newblock Cross-lingual text-to-speech using multi-task learning and speaker
  classifier joint training.
\newblock \emph{arXiv preprint arXiv:2201.08124}.

\bibitem[{Zeghidour et~al.(2021)Zeghidour, Luebs, Omran, Skoglund, and
  Tagliasacchi}]{zeghidour2021soundstream}
Neil Zeghidour, Alejandro Luebs, Ahmed Omran, Jan Skoglund, and Marco
  Tagliasacchi. 2021.
\newblock Soundstream: An end-to-end neural audio codec.
\newblock \emph{IEEE/ACM Transactions on Audio, Speech, and Language
  Processing}, 30:495--507.

\bibitem[{Zhang et~al.(2022)Zhang, Lv, Guo, Shao, Yang, Xie, Xu, Bu, Chen,
  Zeng, Wu, and Peng}]{Zhang_ISCA2022_Wenetspeech}
Binbin Zhang, Hang Lv, Pengcheng Guo, Qijie Shao, Chao Yang, Lei Xie, Xin Xu,
  Hui Bu, Xiaoyu Chen, Chenchen Zeng, Di~Wu, and Zhendong Peng. 2022.
\newblock \href {https://doi.org/10.1109/ICASSP43922.2022.9746682}
  {{WENETSPEECH:} {A} 10000+ hours multi-domain mandarin corpus for speech
  recognition}.
\newblock In \emph{{IEEE} International Conference on Acoustics, Speech and
  Signal Processing, {ICASSP} 2022, Virtual and Singapore, 23-27 May 2022},
  pages 6182--6186. {IEEE}.

\bibitem[{Zhang et~al.(2021)Zhang, Tan, Ren, Qin, Zhang, and
  Liu}]{Zhang_AAAI2021_UWSpeech}
Chen Zhang, Xu~Tan, Yi~Ren, Tao Qin, Kejun Zhang, and Tie{-}Yan Liu. 2021.
\newblock \href {https://ojs.aaai.org/index.php/AAAI/article/view/17684}
  {Uwspeech: Speech to speech translation for unwritten languages}.
\newblock In \emph{Thirty-Fifth {AAAI} Conference on Artificial Intelligence,
  {AAAI} 2021, Thirty-Third Conference on Innovative Applications of Artificial
  Intelligence, {IAAI} 2021, The Eleventh Symposium on Educational Advances in
  Artificial Intelligence, {EAAI} 2021, Virtual Event, February 2-9, 2021},
  pages 14319--14327. {AAAI} Press.

\bibitem[{Zhang et~al.()Zhang, Ye, Ko, Mingxuan, and Yaqian}]{dongzhang2023}
Dong Zhang, Rong Ye, Tom Ko, Wang Mingxuan, and Zhou Yaqian.
\newblock Dub: Discrete unit back-translation for speech translation.
\newblock In \emph{Findings in ACL 2023}.

\bibitem[{Zhang et~al.(2023)Zhang, Zhou, Wang, Chen, Wu, Liu, Chen, Liu, Wang,
  Li et~al.}]{zhang2023speak}
Ziqiang Zhang, Long Zhou, Chengyi Wang, Sanyuan Chen, Yu~Wu, Shujie Liu, Zhuo
  Chen, Yanqing Liu, Huaming Wang, Jinyu Li, et~al. 2023.
\newblock Speak foreign languages with your own voice: Cross-lingual neural
  codec language modeling.
\newblock \emph{arXiv preprint arXiv:2303.03926}.

\end{thebibliography}
\bibliographystyle{acl_natbib}




\end{document}